\documentclass[sigconf]{acmart}
\usepackage{graphicx}
\usepackage{booktabs}
\usepackage{balance}
\usepackage{array}
\usepackage{makecell}
\usepackage{xcolor}

\usepackage{dingbat}
\usepackage{multirow}
\usepackage{tabularx}
\usepackage{colortbl}
\usepackage{pifont}
\usepackage{hyperref}
\AtBeginDocument{%
  }

\copyrightyear{2025}
\acmYear{2025}
\setcopyright{acmlicensed}\acmConference[MM '25]{Proceedings of the 33rd ACM International Conference on Multimedia}{October 27--31, 2025}{Dublin, Ireland}
\acmBooktitle{Proceedings of the 33rd ACM International Conference on Multimedia (MM '25), October 27--31, 2025, Dublin, Ireland}
\acmDOI{10.1145/3746027.3755479}
\acmISBN{979-8-4007-2035-2/2025/10}
\begin{document}

\title{Compute Only 16 Tokens in One Timestep: Accelerating Diffusion Transformers with Cluster-Driven Feature Caching}


\author{Zhixin Zheng}
\authornote{Both authors contributed equally to this research.}
\affiliation{
    \institution{Shanghai Jiao Tong University}
    \city{Shanghai}
    \country{China}
}
\affiliation{
    \institution{University of Electronic Science and Technology of China}
    \city{Chengdu}
    \country{China}
}

\author{Xinyu Wang}
\authornotemark[1]
\affiliation{
    \institution{Shanghai Jiao Tong University}
    \city{Shanghai}
    \country{China}
}
\affiliation{
    \institution{Shandong University}
    \city{Qingdao}
    \country{China}
}

\author{Chang Zou}
\affiliation{
    \institution{Shanghai Jiao Tong University}
    \city{Shanghai}
    \country{China}
}
\affiliation{
    \institution{University of Electronic Science and Technology of China}
    \city{Chengdu}
    \country{China}
}

\author{Shaobo Wang}
\affiliation{
    \institution{Shanghai Jiao Tong University}
    \city{Shanghai}
    \country{China}
}

\author{Linfeng Zhang}
\authornote{Corresponding author.}
\affiliation{
    \institution{Shanghai Jiao Tong University}
    \city{Shanghai}
    \country{China}
}
\email{zhanglinfeng@sjtu.edu.cn}

\renewcommand{\shortauthors}{Zhixin Zheng, Xinyu Wang, Chang Zou, Shaobo Wang, \& Linfeng Zhang}


\begin{abstract}
    Diffusion transformers have gained significant attention in recent years for their ability to generate high-quality images and videos, yet still suffer from a huge computational cost due to their iterative denoising process. 
    Recently, feature caching has been introduced to accelerate diffusion transformers by caching the feature computation in previous timesteps and reusing it in the following timesteps, which leverage the temporal similarity of diffusion models while ignoring the similarity in the spatial dimension.
    In this paper, we introduce Cluster-Driven Feature Caching (ClusCa) as an orthogonal and complementary perspective for previous feature caching. Specifically, ClusCa performs spatial clustering on tokens in each timestep, computes only one token in each cluster and propagates their information to all the other tokens, which is able to reduce the number of tokens by over 90\%.
    Extensive experiments on DiT, FLUX and HunyuanVideo demonstrate its effectiveness in both text-to-image and text-to-video generation. Besides, it can be directly applied to any diffusion transformer without requirements for training.
    For instance, ClusCa achieves 4.96$\times$ acceleration on FLUX with an ImageReward of 99.49\%, surpassing the original model by 0.51\%.
    The code is available at \url{https://github.com/Shenyi-Z/Cache4Diffusion}.

\end{abstract}

\ccsdesc[500]{Computing methodologies~Computer vision}

\keywords{Diffusion Transformer, Acceleration, Feature Caching, Text-to-Image Generation, Text-to-Video Generation}


\maketitle

\section{Introduction}\label{sec:introduction}    
Recent advancements in diffusion-based generative frameworks have revolutionized the field of visual synthesis, achieving unprecedented performance in both image~\cite{DDPM} and video generation~\cite{blattmann2023stablevideodiffusionscaling}. The introduction of transformer-architected diffusion models (DiTs)~\cite{DiT} has particularly enhanced generation capabilities through scalable parameterization, setting new state-of-the-art benchmarks in output fidelity. However, the inherent computational complexity of these architectures poses significant deployment challenges, especially for latency-sensitive applications. This limitation primarily stems from the iterative nature of the denoising process, which requires repeated network evaluations across multiple timesteps - a procedure whose computational cost scales linearly with both model complexity and sampling steps.

\begin{figure}[t]
    \centering\includegraphics[width=\linewidth]{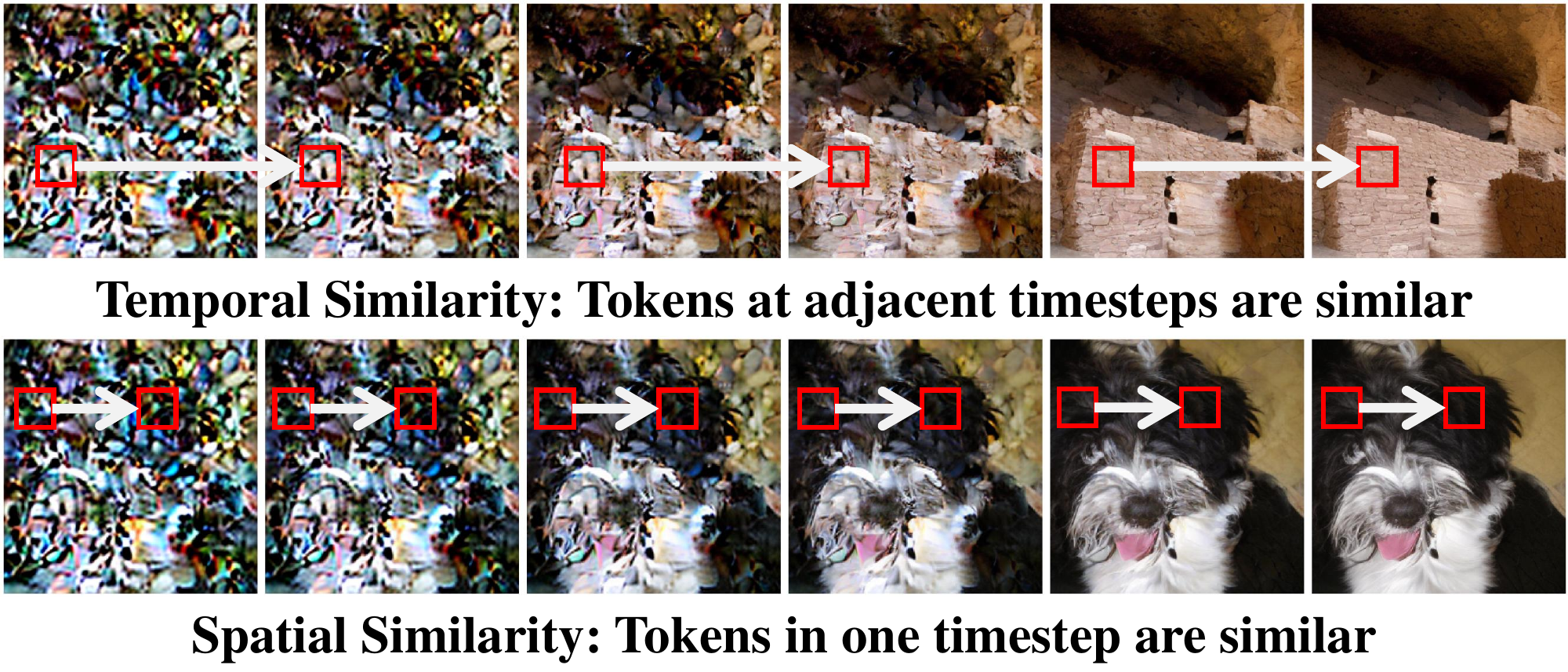}
    \vspace{-7mm}
    \centering\caption{Visualization of similarity in two dimensions}
    \label{fig:sptiotemporal similarity}
\end{figure}

\begin{figure}[t]
    \centering\includegraphics[width=\linewidth]{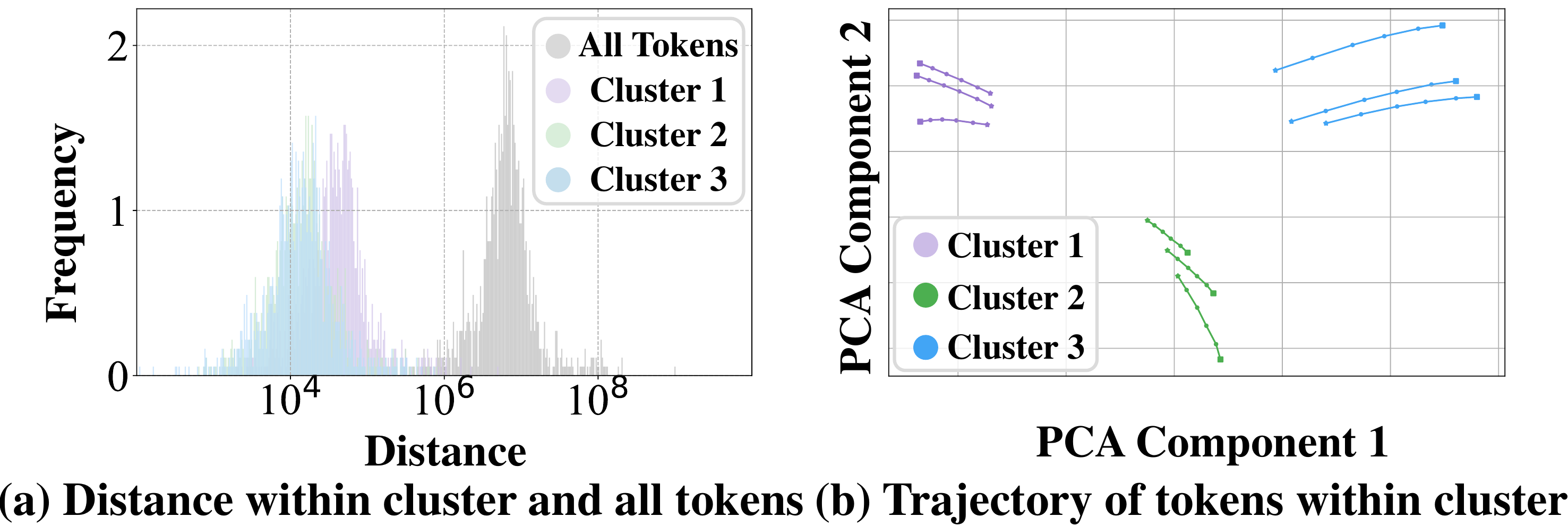}
    \centering\caption{$\quad$(a) The distributions of distance between tokens in the same cluster (1/2/3) and all the tokens in the current timestep, showing that tokens in the same cluster have significantly lower distance.
    (b) PCA visualization on the evolution of tokens in different clusters. Points in the same line denote the same token in different timesteps. Tokens with the same color denote tokens in the same cluster.}
    \label{fig:distance distribution and trajectory}
    \vspace{-4mm}
\end{figure}

To mitigate these efficiency constraints, there are dual optimization pathways: algorithmic improvements for step reduction~\cite{ma2023deepcacheacceleratingdiffusionmodels} and architectural enhancements for network acceleration~\cite{yuan2024ditfastattnattentioncompressiondiffusion, zhao2025realtimevideogenerationpyramid}. Among these strategies, temporal feature cache and reuse mechanisms have gained attention due to their plug-and-play advantages. It is grounded in the empirical observation that features exhibit high similarity between adjacent timesteps during the denoising process. This enables the cache of computed features for direct reuse in subsequent timesteps, thereby allowing the diffusion model to skip computational operations at specific timesteps while maintaining denoising effectiveness.  
Early implementations in U-Net-based frameworks exploit skip connections to recycle intermediate activations across neighboring timesteps~\cite{wimbauer2024cachecanacceleratingdiffusion,ma2023deepcacheacceleratingdiffusionmodels}. As transformer-based diffusion models (DiTs)~\cite{DiT} dominate state-of-the-art benchmarks, many attempts to adapt caching principles to DiTs have yielded success~\cite{FORA, ToCa, DuCa, TaylorSeer}. 
As shown in Figure~\ref{fig:sptiotemporal similarity}, such feature caching methods accelerate the inference process by leveraging the \emph{\textbf{temporal similarity}} of tokens, \emph{i.e.}, the similarity between the token in adjacent timesteps. However, with the distance between the timesteps increasing, the similarity between their features decays significantly, leading to a significant drop in generation under high acceleration ratios.

In this paper, we propose to solve this problem from the perspective of the \emph{\textbf{spatial dimension}}. As shown in Figure~\ref{fig:sptiotemporal similarity}, tokens of the diffusion model in the same timestep also exhibit significantly high similarity. To further study this phenomenon, we cluster the tokens of diffusion models at different spatial positions in the same timestep with K-Means and study their distance. As shown in Figure~\ref{fig:distance distribution and trajectory}(a), the distance between tokens within the same cluster is around 100$\times$ smaller than the distance between all the tokens. Besides, Figure~\ref{fig:distance distribution and trajectory}(b) visualizes the features of nine tokens belonging to three different clusters across adjacent 6 timesteps (plotted in a single line), demonstrating that tokens belonging to the same cluster exhibit different variations during the denoising timesteps. These two observations demonstrate the similarity between tokens within the same cluster in both their features and the variations on their features, indicating that \emph{it is possible to compute only one token in each cluster and then reuse its features for the remaining tokens in the same cluster.}




\begin{figure}[t]
    \centering\includegraphics[width=\linewidth]{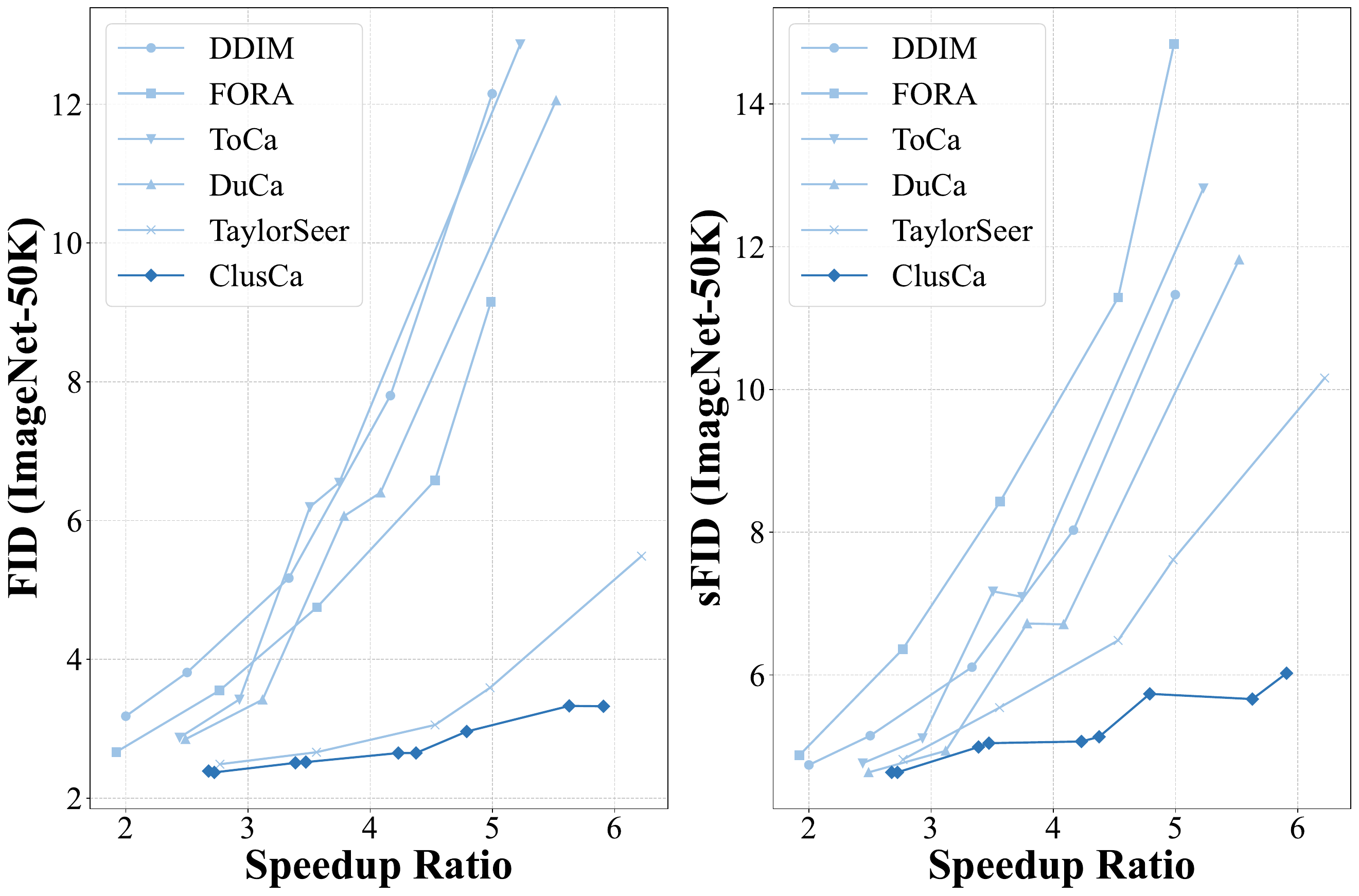}
    \vspace{-6mm}
    \centering\caption{Comparison between existing feature caching methods and ClusCa on DiT with FID and sFID (lower is better).}
    \label{fig:FID_sFID_performance}
    \vspace{-6mm}
\end{figure}

Based on this observation, this paper introduces Cluster-Driven Feature Caching (\textit{ClusCa}) that takes advantage of both the temporal and spatial similarity of tokens. Specifically, we uniformly divide all the timesteps into several cache cycles. (I) At the first timestep of each cache cycle, we compute all the tokens (full calculation), store them in the cache, and cluster these tokens with K-Means based on their values in the last layer. (II) In the following timesteps of this cache cycle, we only compute one token per cluster (partial calculation). (III) Then, for the left tokens in each cluster, we update their value with a weighted sum between their values in the previous timestep (\emph{i.e.,} temporal reuse) and the computed token in their corresponding cluster (\emph{i.e.,} spatial reuse). 
Such a dual feature reusing strategy allows \textit{ClusCa} to leverage both spatial and temporal similarity to approximate the uncomputed tokens, therefore allowing us to achieve a significant acceleration without a drop in generation quality. Note that the ratio of computational costs for clustering is usually $\leq5\%$ since we perform K-Means in the last layer of the full-calculation timestep.

Extensive experiments have been conducted in DiT, FLUX, and HunyuanVideo to demonstrate their effectiveness.
As shown in Figure~\ref{fig:FID_sFID_performance}, benefiting from spatial feature reusing, it achieves significantly lower FID in high acceleration ratios. On FLUX and HunyuanVideo, \textit{ClusCa} achieves around 4.96$\times$ and 6.21$\times$ acceleration while maintaining the generation quality.

In summary, our contributions are as follows. 
\begin{itemize}
\item \textbf{Spatial Similarity Analysis in Diffusion Transformers: } Through systematic investigation of spatial redundancies in DiTs, we reveal the similarity in both features of tokens in the same cluster and the cluster formations remain stable over timesteps. 
\item \textbf{Spatiotemporal Feature Reuse Paradigm: } We introduce \textit{ClusCa}, an innovative approach that achieves full token updates through the computation of only a few tokens from distinct clusters, reducing computational complexity while mitigating the error accumulation caused by caching. 
\item \textbf{State-of-the-Art Performance: } Through comprehensive benchmarking on mainstream diffusion architectures including DiT and FLUX for image synthesis tasks, our method demonstrates superior efficiency-quality trade-offs. It provides a solution to obtain higher generation quality under high speedup ratio. 
\end{itemize}
\vspace{-3mm}
\begin{figure*}[t]
    \centering
    \includegraphics[width=0.98\linewidth]{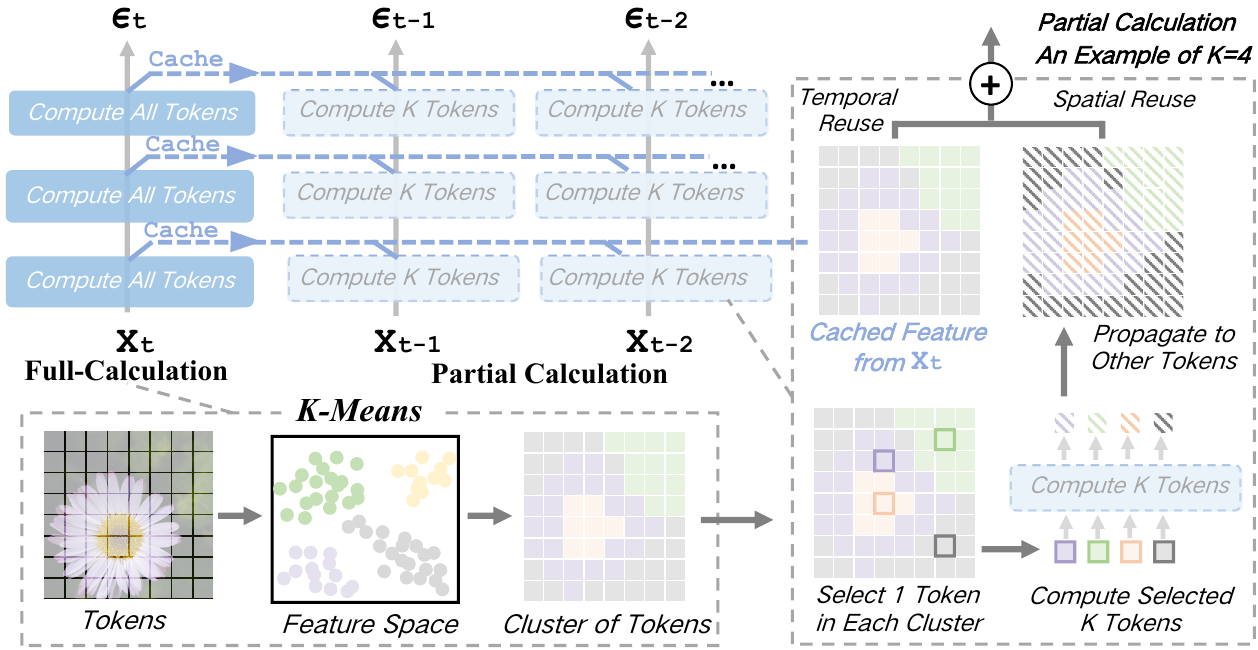}
    \vspace{-2mm}
    \caption{The Overview of ClusCa:$\quad$(a) In full-calculation step, ClusCa caches tokens and reuses them across timesteps by leveraging temporal similarity, enabling partial calculation for improved efficiency. In the partial-calculation step, ClusCa further exploits spatial similarity in the feature space by clustering tokens based on feature distance and reusing shared information within local regions.
    $\quad$(b) The spatial reuse process includes token clustering, selective update, temporal reuse from cache, and cluster-wise token merging to reduce redundant computation while preserving representation quality.}
    \label{fig:method}
    \vspace{-4mm}
\end{figure*}

\section{Related Works}\label{sec:related_works}

The recent advancements in generative models have witnessed diffusion-based approaches emerging as a dominant paradigm for content synthesis~\cite{DDPM}, demonstrating exceptional capabilities in image and video generation that surpass previous state-of-the-art generative architectures~\cite{dhariwal2021diffusionmodelsbeatgans}. A pivotal development occurred with the introduction of diffusion transformers, which effectively addressed the scalability limitations inherent in conventional U-Net architectures, enabling diffusion models to achieve unprecedented performance across diverse vision tasks, including but not limited to image generation, inpainting, and super-resolution. However, the diffusion models are still limited by their multi-step iterative generation mechanism, which brings a heavy computational burden. To mitigate this efficiency bottleneck, many works have been carried out to optimizing diffusion frameworks in two primary directions: (1) reducing the number of diffusion steps, and (2) enhancing the network efficiency.

\vspace{-4mm}

\subsection{Temporal Step Optimization}

In order to maintain high-quality generation while reducing the number of diffusion steps, DDIM~\cite{DDIM} breaks through the iteration constraints of traditional markov chains and establishes a theoretical framework for deterministic sampling process, achieving order-of-magnitude reductions in iteration steps. DPM-Solver family~\cite{dpmsolver, dpmsolver++} significantly improved convergence rates by modeling higher-order ODE solvers. From an optimal transport perspective, Rectified Flow~\cite{liu2022flowstraightfastlearning} systematically straightens non-linear probabilistic flow trajectories via optimally guided deterministic ODEs, enabling high-quality sample generation with significantly fewer inference steps. Consistency models~\cite{song2023consistencymodels} enforce self-consistency constraints to directly map data from noise to the target distribution through a deterministic process, achieving high-quality sample generation in single or few steps while maintaining competitive sample quality.

\vspace{-2mm}

\subsection{Network Efficiency Enhancement}

\noindent \textbf{Architectural Optimization. } Nowadays, substantial progress has been made in addressing Diffusion Transformer's computational bottlenecks through parameter reduction strategies. Parameter Slimming methods employ three principal compression paradigms: (1) structured pruning~\cite{fang2023structuralpruningdiffusionmodels, zhu2024dipgodiffusionprunerfewstep} removes redundant weights; (2) quantization~\cite{kim2025dittoacceleratingdiffusionmodel, li2023qdiffusionquantizingdiffusionmodels, shang2023posttrainingquantizationdiffusionmodels} enables low-bit computations; (3) token compression~\cite{bolya2023tokenmergingfaststable, cheng2025catpruningclusterawaretoken, kim2023tokenfusionbridginggap, saghatchian2025cachedadaptivetokenmerging, zhang2024tokenpruningcachingbetter, zhang2025sito} dynamically shortens sequence lengths. Such methods inevitably lead to information loss, necessitating delicate fine-tuning compensation mechanisms.

\noindent \textbf{Feature Caching Paradigms. } Recent advancements in temporal redundancy utilization also demonstrate significant acceleration potential. DeepCache~\cite{ma2023deepcacheacceleratingdiffusionmodels} first observes the significant temporal similarity in high-level features in denoising process and proposes a caching mechanism to store and reuse high-level features across different diffusion steps, reducing redundant calculations significantly while achieving superior efficacy than pruning and distillation algorithms without additional training as they do. Subsequent works have extended this concept to diffusion transformers. For example, FORA~\cite{FORA} and $\Delta$-DiT~\cite{chen2024deltadittrainingfreeaccelerationmethod} leverage cached features and residual of features to accelerate the diffusion process, ToCa~\cite{ToCa} and DuCa~\cite{DuCa} implement token-wise feature caching strategy to fresh cache during each caching cycle. Recently, TaylorSeer~\cite{liu2025reusingforecastingacceleratingdiffusion} converts the paradigm "cache-then-reuse" into "cache-then-forecast" by utilizing Taylor series expansion to approximate the trajectory of features at future timesteps, obtaining high acceleration with negligible quality degradation. 


\section{Method}\label{sec:methodology}

\begin{figure*}[t]
    \centering\includegraphics[width=0.9\linewidth]{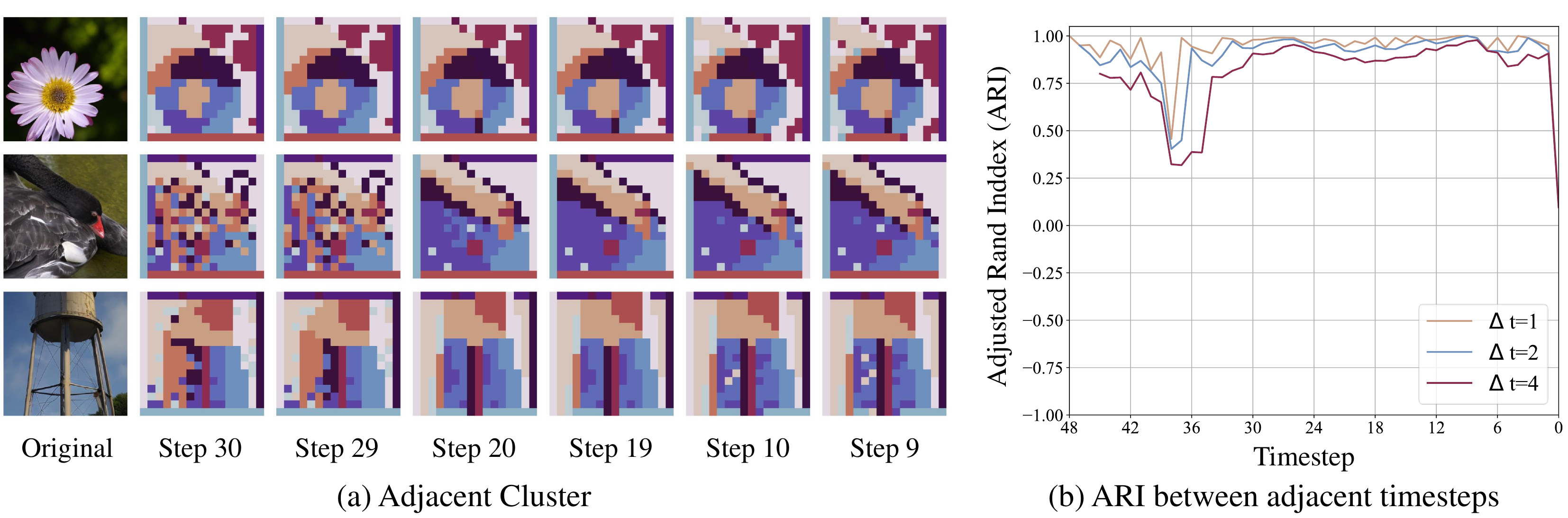}
    \vspace{-4mm}
    \centering\caption{$\quad$(a) Visualization of clustering results in adjacent steps. $\quad$(b) Adjusted rand index (ARI) of clusters in adjacent steps, where $\Delta t$ is the timestep span between compared clusters. A higher ARI indicates the adjacent timesteps have closer clustering results. For most timesteps, the ARI exceeds 0.8, indicating high similarity, or even identity, between adjacent clusters.} 
    \label{fig:similarity_cluster}
    \vspace{-4mm}
\end{figure*}

\subsection{Preliminary}

\noindent \textbf{Diffusion Models. } Diffusion models are a series of image generative models, which generally include the forward and backward process. The forward process transforms clean image from the prior data distribution to noise by gradually adding random Gaussian noise:
$$
x_t=\sqrt{\alpha_t}x_0+\sqrt{1-\alpha_t}\epsilon_t
$$
where $\epsilon_t$ denotes the noise added in $t$-th timesteps. The noise schedule $\{\alpha_t\}_{t=1}^T$ is designed to decrease monotonically with $t$, ensuring noise is smoothly added into the clean image. After $T$ timesteps processing, the final data $x_T$ becomes pure Gaussian noise.

The backward process aims to learn to reverse noise by gradually denoising the data $x_t$. Specifically, at each step, the denoising operation can be formulated as:
$$
x_{t-1}=\frac{1}{\sqrt{\alpha_t}}\left(x_t-\frac{1-\alpha_t}{\sqrt{1-\alpha_t}} \epsilon_\theta\left(x_t, t\right)\right)+\sigma_t \epsilon
$$
where $\epsilon_\theta(x_t,t)$ is the neural network model trained to predict the noise $\epsilon_t$ at each timestep. $\epsilon \sim \mathcal N(0,I)$ is additional noise controlled by $\sigma_t$ in reverse process. In the past, the U-Net architecture has demonstrated remarkable effectiveness in modeling the denoising function $\epsilon_\theta$. Recently, pure transformer-based architecture has been proposed instead U-Net for modeling $\epsilon_\theta$, which shows superior generative capabilities.

\noindent \textbf{Diffusion Transformer. }
The Diffusion Transformer (DiT) is a pure transformer-based architecture designed for diffusion models, characterized by its fully transformer-based backbone and adaptive normalization mechanisms. It processes a sequence of tokens $\mathrm{x}_t=\{ x_i \}_{i=1}^{H\times W}$, where each token $x_i$ corrsponds a patch of the image. The model comprises a sequential stack of transformer blocks $\mathcal{G} = g_1 \circ g_2 \circ \cdots \circ g_L$, each one integrates three core components: self-attention (SA), cross-attention (CA) and multilayer perceptron (MLP), formally expressed as $g_l = \{f_{\text{SA}}^l, f_{\text{CS}}^l, f_{\text{MLP}}^l \}$. 

\noindent \textbf{Feature Caching for Diffusion Transformer. } 
Existing feature caching techniques for diffusion transformer~\cite{FORA} adopt a \textit{periodic} temporal caching strategy with interval $\mathcal{N}$. During a period of adjacent timesteps $\{t, t-1, t-2, \dots, t-\mathcal{N}+1\}$, the calculation of the feature in $t$-th timestep will be cached through the mapping: $\mathcal{C}(x_t^l):= \mathcal{F}(x_t^l)$, for $\forall l \in [0, L)$, and then reused in the subsequent $\mathcal{N}-1$ timesteps to skip computation via: $\mathcal{F}(x_{t-k}^l):=\mathcal{C}(x_t^l)$, for $\forall k \in [1, \mathcal{N})$, reducing FLOPs by approximately $\frac{\mathcal{N}-1}{\mathcal{N}}$. In order to reduce the error accumulation as $\mathcal{N}$ increases, ToCa~\cite{ToCa} introduces a more  fine-grained caching method in the token-level: during skip timesteps, only important tokens will be selected to update cache, which can be formulated as:
$$
\mathcal{C}(x_i):=\mathcal{F}(x_i)\ \text{for}\ i\in\mathcal{I}_\text{Compute},
$$
where $\mathcal{I}_\text{Compute}$ is selected based on a score mechanism. 

\noindent \textbf{TaylorSeer}
~\cite{TaylorSeer} introduces Taylor series-based predictive caching to mitigate error accumulation in conventional caching methods. Instead of directly caching features, it defines a cache storing the feature and its $m$-th order temporal differences, that is:
$$
\mathcal{C}(x_t^l):=\{\mathcal{F}(x_t^l),\Delta\mathcal{F}(x_t^l),\dots,\Delta^m\mathcal{F}(x_t^l)\}
$$
where $\Delta^i\mathcal{F}(x_t^l)$ is the $i$-th order difference. Then, features in skip timesteps can be approximately predicted as follows:
$$
\mathcal{F}_{\mathrm{pred},m}(x_{t-k}^l)=\mathcal{F}(x_t^l)+\sum_{i=1}^m\frac{\Delta^i\mathcal{F}(x_t^l)}{i!\cdot N^i}(-k)^i.
$$
Our proposed \textit{ClusCa} is also based on this forecast mechanism. 

\subsection{Spatial Similarity in Feature}


While conventional feature caching approaches predominantly focus on temporal coherence of individual tokens across adjacent timesteps, we conduct a systematic investigation into the spatial similarity patterns among tokens during the denoising trajectory.

\noindent \textbf{Observation 1. }\label{observation 1} \textit{Tokens in the same cluster demonstrate not only significant feature similarity but also exhibit high consistency in motion patterns. }

In Figure~\ref{fig:distance distribution and trajectory}, we provide robust empirical evidence to support this observation. We conducted comparative analyses on the intra-cluster token distances versus global token distances across multiple feature dimensions. 
A fundamental property of clustering is that intra-cluster distances are significantly smaller than global distances(Figure~\ref{fig:distance distribution and trajectory} a).
In addition, through dimensionality reduction via principal component analysis (PCA), we projected feature trajectories of tokens across sequential time steps into a 2-dimensional latent space. Visualization of these trajectories~\ref{fig:distance distribution and trajectory} (b) demonstrates that tokens belonging to the same cluster exhibit tightly aligned evolutionary paths with directional coherence, while maintaining significantly smaller inter-trajectory distances compared to cross-cluster pairs. This geometric regularity reveals that the clustering mechanism captures not only static feature similarity but also dynamic feature covariation patterns throughout temporal evolution processes.

\noindent \textbf{Observation 2. }\label{observation 2} \textit{The cluster formations maintain continuous morphological similarity across adjacent time steps. } 

Figure~\ref{fig:similarity_cluster} demonstrates substantial similarity of clustering formations across consecutive temporal intervals. To quantitatively assess this temporal coherence, we implemented a systematic evaluation framework employing the adjusted Rand index (ARI) - a statistical measure ranging from $[-1,1]$ where values approaching 1 indicate perfect cluster correspondence. Our analysis reveals sustained ARI values maintaining above 0.8 across most of timesteps, demonstrating temporal consistency in cluster formations, also eliminating the need for frequent clustering, but only at the beginning of the cache cycle or other critical steps to minimize the additional overhead of clustering. Specifically, our experimental results demonstrate that the computation costs for clustering in \textit{ClusCa} are usually $\leq$ 5\%.

\subsection{Cluster-Driven Feature Caching}
Previous feature caching methods treat all tokens equally or update a few tokens according to their importance, neglecting the intrinsic connections between tokens. Tokens in the latent feature space exhibit a spatial structure similar to image pixels, where tokens within local regions are often highly correlated due to the continuity of visual content. To effectively extract similarity relationships in input tokens, the K-Means algorithm is chosen for clustering because of its simplicity and high efficiency. Specifically, K-Means performs clustering with input $\{x_i\}^{H\times W}_{i=1}$, where $x_i$ is a $d$-dimensional real vector corresponding to the feature of each token, and returns a cluster index vector $\mathbf{I}\in \mathbb{R}^{H\times W}$, where the elements $\mathbf{I}_i$ satisfy $0\le \mathbf{I}_i < K$, indicating the cluster number to which the $i$-th token belongs. As K-Means aims to minimize the pairwise deviations of points in the same cluster:
$$
\arg\min_{S}\sum_{i=1}^K\frac{1}{|S_i|}\sum_{\mathbf{x},\mathbf{y}\in S_i}\|\mathbf{x}-\mathbf{y}\|^2,
$$
where $S=\{S_1,S_2,\dots,S_K\}$ denotes a partition of tokens, the distance between tokens in the same cluster is much smaller than that of all tokens as mentioned before. 

In order to avoid too much extra overhead caused by clustering, based on Observation \ref{observation 2}, we only perform clustering at full-calculation timesteps and cluster centroids will be cached at the same time, which can be used for accelerating the convergence of future clustering as there will be no need for initialize cluster centroids randomly. The time consumption analysis of clustering can be found in the supplementary material. 





Based on the clustering results, we compute only a subset of tokens selected from each cluster. Random selection is applied in order to ensure broad spatial representation. Empirically, we observe that \emph{selecting a single token per cluster already achieves strong performance while substantially reducing computational complexity.} To optimize the speedup ratio, we therefore fix the number of selected tokens per cluster to \emph{one} throughout our implementation. 

\subsection{Spatial Feature Reuse}

Once representative tokens are selected and recalculated, the next key challenge is how to propagate their updated feature information back to the other tokens within the same cluster to avoid redundant computations.

Intuitively, since tokens within a cluster are semantically similar, we can reuse the updated features of the computed representatives to guide the cache update of non-recomputed tokens in the same cluster. In this way, we are able to balance efficiency and accuracy: computing only a small number of tokens while still maintaining informative feature representations for all tokens.
Specifically, the feature cache of recomputed tokens is directly updated with their newly calculated values, which is straightforward. For non-recomputed tokens, their cache updates comprise two components: one inherited from their previous cache storage, and another derived from the recomputed values of co-clustered tokens. This can be formally formulated as follows:
$$
\mathcal{C}(x_i)=\begin{cases}\mathcal{F}(x_i),&i\in\mathcal{I}_\mathrm{Compute}\\\gamma\cdot\mu_{(i)}+(1-\gamma)\mathcal{C}(x_i),&i\notin\mathcal{I}_\mathrm{Compute}\end{cases}
$$
where $\gamma$ is hyperparameter employed in this weighted summation, we formally term this coefficient the \textbf{propagation ratio}, as it governs how the latest information from co-clustered newly computed tokens is propagated to non-recomputed tokens smoothly; $\mu_{(i)}$ is the mean value of computed feature of tokens in $\mathcal{L}_{\text{compute}}$ for each cluster, formally expressed as:
$$
\mu_{(i)}=\frac{\sum_{j\in\mathcal{I}_{\mathrm{Compute}}}\mathcal{F}(x_j)}{\sum_{j\in\mathcal{I}_{\mathrm{Compute}}}[\mathbf{I}_j=\mathbf{I}_i]},
$$
here, $[\cdot]$ denotes the Iverson bracket. 

It should be noted that selecting the \textbf{ propagation ratio} $\gamma$ involves a critical trade-off. Excessively large values would allow information from co-clustered computed tokens to dominate, thereby discarding historical cache and degenerating the approach into a token-pruning paradigm. Conversely, overly small values would inadequately leverage co-clustered computed information, essentially reducing the method to ToCa~\cite{ToCa} scheme. Empirical observations reveal the quality of the generation exhibits a pronounced trend of initial improvement followed by degradation as $\gamma$ increases, which will be clearly demonstrated in the next section. 

\section{Experiments}\label{sec:experiments}

\begin{table*}[ht]
    \centering
    \caption{\textbf{Quantitative comparison in text-to-image generation} for FLUX on Image Reward.
    }
    \vspace{-3mm}
    \setlength\tabcolsep{7.0pt} 
      \small
      \resizebox{0.90\textwidth}{!}{
      \begin{tabular}{l | c | c  c | c  c | c | c}
        \toprule
        {\bf Method} & {\bf Efficient} &\multicolumn{4}{c|}{\bf Acceleration} &{\bf Image Reward $\uparrow$} &\bf CLIP$\uparrow$ \\
        \cline{3-6}
        {\bf FLUX.1\citep{flux2024}} & {\bf Attention \citep{dao2022flashattentionfastmemoryefficientexact}} & {\bf Latency(s) $\downarrow$} & {\bf Speed $\uparrow$} & {\bf FLOPs(T) $\downarrow$}  & {\bf Speed $\uparrow$} & \bf DrawBench &\bf Score \\
        \midrule
      
      $\textbf{[dev]: 50 steps}$     
                               & \ding{52}  &  {25.82}  & {1.00$\times$} & {3719.50}   & {1.00$\times$} & {0.9898}  &{19.761}      \\ 
      \midrule
    
      {$60\%$\textbf{ steps}}  & \ding{52}  &  {16.70}  & {1.55$\times$} & {2231.70}   & {1.67$\times$} & {0.9663}  &{19.526}      \\
      {$50\%$\textbf{ steps}}  & \ding{52}  &  {13.14}   & {1.96$\times$} & {1859.75}  & {2.00$\times$} & {0.9595}  &{19.455}             \\
      {$40\%$\textbf{ steps}}  & \ding{52}  &  {10.59}   & {2.44$\times$} & {1487.80}   & {2.62$\times$} & {0.9554}  &{19.003}             \\
      {$\Delta$-DiT} ($\mathcal{N}=2$) & \ding{52}  &  {17.80}  & {1.45$\times$} & {2480.01}   & {1.50$\times$} & {0.9444}  &{19.396}      \\
      {$\Delta$-DiT} ($\mathcal{N}=3$) & \ding{52}  &  {13.02}  & {1.98$\times$} & {1686.76}   & {2.21$\times$} & {0.8721}  &{18.742}      \\
      \midrule
      {$34\%$\textbf{ steps}}  & \ding{52}  &  {9.07}   & {2.85$\times$} & {1264.63}   & {3.13$\times$} & {0.9453}  &{18.870}             \\
      
      $\textbf{FORA}$ $(\mathcal{N}=3)$
      \citep{FORA}& \ding{52}  &  {10.16}   & {2.54$\times$} & {1320.07}   & {2.82$\times$} & {0.9776}    &{19.339}   \\
    

      $\textbf{\texttt{ToCa}}$ $(\mathcal{N}=6)$
      \citep{ToCa} 
                               & \ding{56}  &  {13.16}   & {1.96$\times$} & {924.30}   & {4.02$\times$} & {0.9802}   &{18.688}  \\
    
      $\textbf{\texttt{DuCa}} (\mathcal{N}=5)$      
      \citep{DuCa}
                               & \ding{52}  &  {8.18}    & {3.15$\times$} & {978.76}   & {3.80$\times$} &\underline{0.9955}   &{19.314}  \\

      $\textbf{TaylorSeer} $ $(\mathcal{N}=4,O=2)$ 
                               & \ding{52}  &  {9.24}    &  {2.80}{$\times$} &  {{1042.27}}  &  {{3.57}$\times$} &{0.9857}   &\bf{19.496}  \\

        \rowcolor{gray!20}
      $\textbf{ClusCa} $ $(\mathcal{N}=5,O=1,K=16)$ 
                               & \ding{52}  &  {8.12}  &  {3.18}{$\times$}  &  \bf{{897.03}}  &  \bf{4.14$\times$}   &{0.9825}   &\underline{19.481}  \\
      \rowcolor{gray!20}
      $\textbf{ClusCa} $ $(\mathcal{N}=5,O=2,K=16)$ 
                               & \ding{52}  &  {8.19}  &  {3.15}{$\times$}  &  {{897.03}}  &  {4.14$\times$}   &\textbf{0.9961}   &{19.422}  \\
      \midrule
      
    
      {$22\%$\textbf{ steps}}  & \ding{52}  &  {6.04}   & {4.28$\times$} & {818.29}   & {4.55$\times$} & {0.8183}  &{18.224}             \\
      
      $\textbf{FORA}$ $(\mathcal{N}=4)$ 
      \citep{FORA}& \ding{52}  &  {8.12}   & {3.14$\times$} & {967.91}   & {3.84$\times$} & {0.9730}    &{19.210}   \\

      
      $\textbf{\texttt{ToCa}}$ $(\mathcal{N}=8)$  
      \citep{ToCa} 
                               & \ding{56}  &  {11.36}   & {2.27$\times$} & {784.54}   & {4.74$\times$} & {0.9451}   &{18.402}  \\
    $\textbf{\texttt{DuCa}}$ $(\mathcal{N}=7)$  
      \citep{DuCa}
                               & \ding{52}  &  {6.74}    & {3.83$\times$} & {760.14}   & {4.89$\times$} &{0.9757}   &{18.962}  \\   

          \textbf{TeaCache} $({l}=0.8)$   \citep{liu2024timestep}& \ding{52}  & 7.21 & 3.58$\times$ & 892.35 & 4.17$\times$  &  0.8683  & 18.500\\ 
          
      $\textbf{TaylorSeer} $ $(\mathcal{N}=5,O=2)$ 
                               & \ding{52}  &  {7.46}    &  {3.46}{$\times$} &  {{893.54}}  &  {{4.16}$\times$} &\underline{0.9864}   &{19.406}  \\
    

      \rowcolor{gray!20}
      $\textbf{ClusCa} $ $(\mathcal{N}=6,O=1,K=16)$ 
                              & \ding{52}  &  {7.10}    & {3.63$\times$} & \bf{748.48}  & \bf{4.96$\times$} &{0.9762}   &\textbf{19.533}  \\
      \rowcolor{gray!20}
      $\textbf{ClusCa} $ $(\mathcal{N}=6,O=2,K=16)$ 
                              & \ding{52}  &  {7.12}  & {3.62$\times$}  & {748.48}   & {4.96$\times$} & \bf{0.9949}   &\underline{19.453}  \\
    
        \bottomrule
      \end{tabular}}
      
      \label{table:FLUX-Metrics}
    \end{table*}

\subsection{Experiment Settings}
\noindent \textbf{Model Configurations.}
We have evaluated the effectiveness of our proposed method on several commonly-used vision generation models: the class-conditional image generation model \textbf{DiT-XL/2}~\cite{DiT}, the
text-to-image generation model \textbf{FLUX.1-dev}~\cite{flux2024}, the text-to-video generation model \textbf{Hunyuan-Video}~\cite{sun2024hunyuanlargeopensourcemoemodel}. The model configuration can be found in the supplementary material. 

\noindent \textbf{Evaluation and Metrics. } 
For class-conditional image generation, we uniformly sample images from all 1,000 ImageNet~\cite{ImageNet} categories, producing 50,000 images at $256\times256$ resolution. Our primary evaluation metric is FID-50k~\cite{FID-50K}, with sFID serving as complementary measure for comprehensive assessment. For text-to-image generation, we employed 200 DrawBench~\cite{saharia2022photorealistictexttoimagediffusionmodels} prompts as the inference input and evaluated the quality of generated images through Image Reward~\cite{xu2023imagerewardlearningevaluatinghuman}, while assessing text-image alignment using CLIP score~\cite{hessel2022clipscorereferencefreeevaluationmetric}. In text-to-video generation tasks, we leveraged the VBench~\cite{huang2023vbenchcomprehensivebenchmarksuite} evaluation framework with 946 distinct prompts, generating five videos per prompt using different random seeds. 

\subsection{Results on Text-to-Image Generation}
\begin{figure}
    \centering
    \includegraphics[width=1\linewidth]{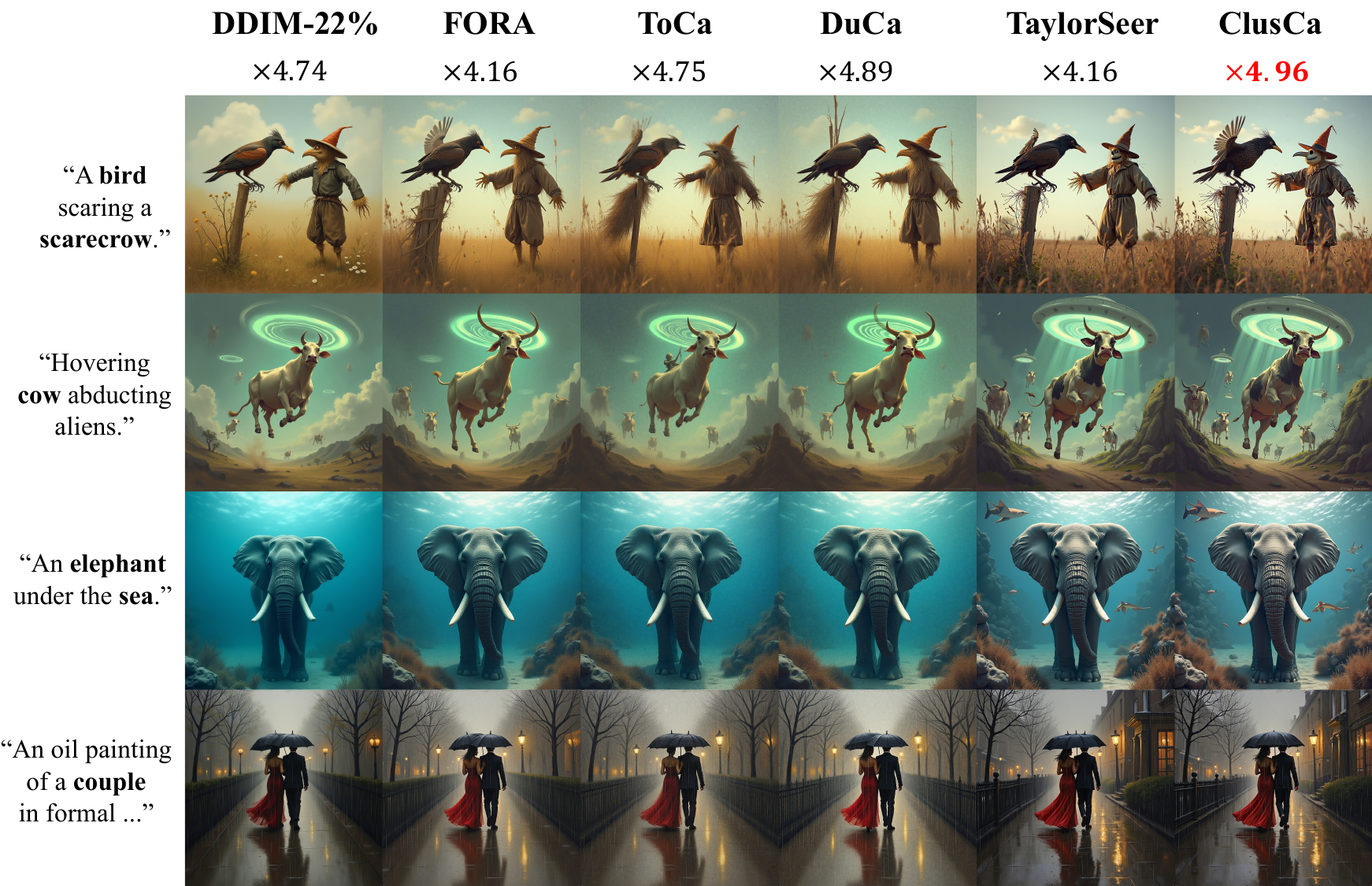}
    \caption{Qualitative comparison on FLUX. ClusCa excels in generating complex scenes and produces images with significantly richer content.}
    \label{fig:FLUX}
    \vspace{-6mm}
\end{figure}

When comparing \textit{ClusCa} with existing feature caching methods, our experimental results demonstrate significant advantages in both efficiency and quality. As shown in Table~\ref{table:FLUX-Metrics}, when $\mathcal{N}=5$, the Image Reward metrics of ToCa and DuCa decrease to 0.9731 and 0.9896 respectively, while \textit{ClusCa} maintains superior generation quality with an Image Reward of 0.9961 while achieving a comparable speedup ratio to TaylorSeer (4.12$\times$ vs. 4.16$\times$). Furthermore, when the speedup ratio exceeds 4.5$\times$, \textit{ClusCa} achieves a higher Image Reward (0.9949) with only marginal latency increase compared to TaylorSeer, demonstrating remarkable performance preservation under extreme acceleration conditions. \textbf{Notably, \textit{ClusCa}'s results on $\mathcal{N}=6$ and $O=2$ with $4.94\times$ speedup ratio (achieving 0.9949 Image Reward) even surpass TaylorSeer's performance on $\mathcal{N}=5$ and $O=2$ with $4.12\times$ (0.9897 Image Reward)}, highlighting our method's superior trade-off advantages in balancing inference speed and generation quality. 
Qualitative comparison with other methods in Figure~\ref{fig:FLUX} highlights \textit{ClusCa}'s superior ability on generating images at high speedup ratio. The fourth case fully proves this, only \textit{ClusCa} successfully produces an umbrella preserving geometric fidelity at 4.96$\times$ acceleration. 

\subsection{Results on Text-to-Video Generation}
\begin{table*}[htbp]
\centering
\caption{\textbf{Quantitative comparison in text-to-video generation} on VBench.}
\vspace{-3mm}
\setlength\tabcolsep{7.0pt} 
\small
\resizebox{0.95\linewidth}{!}{
\begin{tabular}{l | c | c  c | c  c | c }
    \toprule
    {\bf Method} & {\bf Efficient} & \multicolumn{4}{c|}{\bf Acceleration} & {\bf VBench $\uparrow$} \\
    \cline{3-6}
    {} & {\bf Attention} & {\bf Latency(s) $\downarrow$} & {\bf Speed $\uparrow$} & {\bf FLOPs(T) $\downarrow$} & {\bf Speed $\uparrow$} & \bf Score(\%) \\
    \midrule
  
    $\textbf{Original}$ 
                           & \ding{52}  & {334.96} & {1.00$\times$} & {29773.0} & {1.00$\times$} & {80.66} \\ 
    \midrule

    $\textbf{DDIM-22\%}$  
                           & \ding{52}  & {87.01} & {3.85$\times$} & {6550.1} & {4.55$\times$} & {78.74} \\
    $\textbf{FORA}$ $(\mathcal{N}=5)$ 
                           & \ding{52}  & {83.78} & {4.00$\times$} & {5960.4} & {5.00$\times$} & {78.83} \\

    $\textbf{\texttt{ToCa}}$ $(\mathcal{N}=5)$ 
                           & \ding{56}  & {93.80} & {3.57$\times$} & {7006.2} & {4.25$\times$} & {78.86} \\

    $\textbf{\texttt{DuCa}}$ $(\mathcal{N}=5)$ 
                           & \ding{52}  & {87.48} & {3.83$\times$} & {6483.2} & {4.62$\times$} & {78.72} \\
    \textbf{TeaCache}  $({l}=0.4)$       & \ding{52}  & 70.43 & 4.76$\times$ & 6550.1 & 4.55$\times$ & 79.36 \\
    
    \textbf{TeaCache}   $({l}=0.5)$    & \ding{52} & 61.47 & 5.45$\times$ & 5359.1 & 5.56$\times$ & 78.32 \\
    $\textbf{TaylorSeer}$ $(\mathcal{N}=5,O=1)$ 
                           & \ding{52}  & {85.93} & {3.90$\times$} & {5960.4} & {5.00$\times$} & {79.93} \\
    
    $\textbf{TaylorSeer}$ $(\mathcal{N}=6,O=1)$ 
                           & \ding{52}  & {79.46} & {4.22$\times$} & {5359.1} & {5.56$\times$} & {79.78} \\

    \rowcolor{gray!20}
    $\textbf{ClusCa}$ $(\mathcal{N}=5,K=32)$ 
                           & \ding{52}  & {87.35} & {3.83$\times$} & {5968.1} & {4.99$\times$} & {\textbf{79.99}} \\
    
    \rowcolor{gray!20}
    $\textbf{ClusCa}$ $(\mathcal{N}=6,K=32)$ 
                           & \ding{52}  & {81.64} & {4.10$\times$} & {5373.0} & {{5.54}$\times$} & \underline{79.96} \\
    \rowcolor{gray!20}
    $\textbf{ClusCa}$ $(\mathcal{N}=7,K=32)$ 
                           & \ding{52}  & {74.88} & {4.47$\times$} & \bf{4796.2} & {\textbf{6.21}$\times$} & {79.60} \\

    \bottomrule
\end{tabular}}
\label{table:HunyuanVideo-Metrics}
\end{table*}

On Text-to-Video Task, \textit{ClusCa} with $\mathcal{N}=6$ and $K=32$ reduces inference latency to $81.64$ seconds and computational cost to $5373.0$ TFLOPs ($5.52\times$ speedup), achieving a $79.96\%$ VBench score that outperforms all Feature-Cache Methods and Feature-Prediction Methods. With $\mathcal{N}=7$ and $K=64$, performance improves further to $6.21\times$ speedup while maintaining a $79.60\%$ score. These results show that \textit{ClusCa} preserves temporal consistency and reduces computational demands, significantly outperforming previous methods. 

\begin{figure*}[t]
    \centering\includegraphics[width=0.95\linewidth]{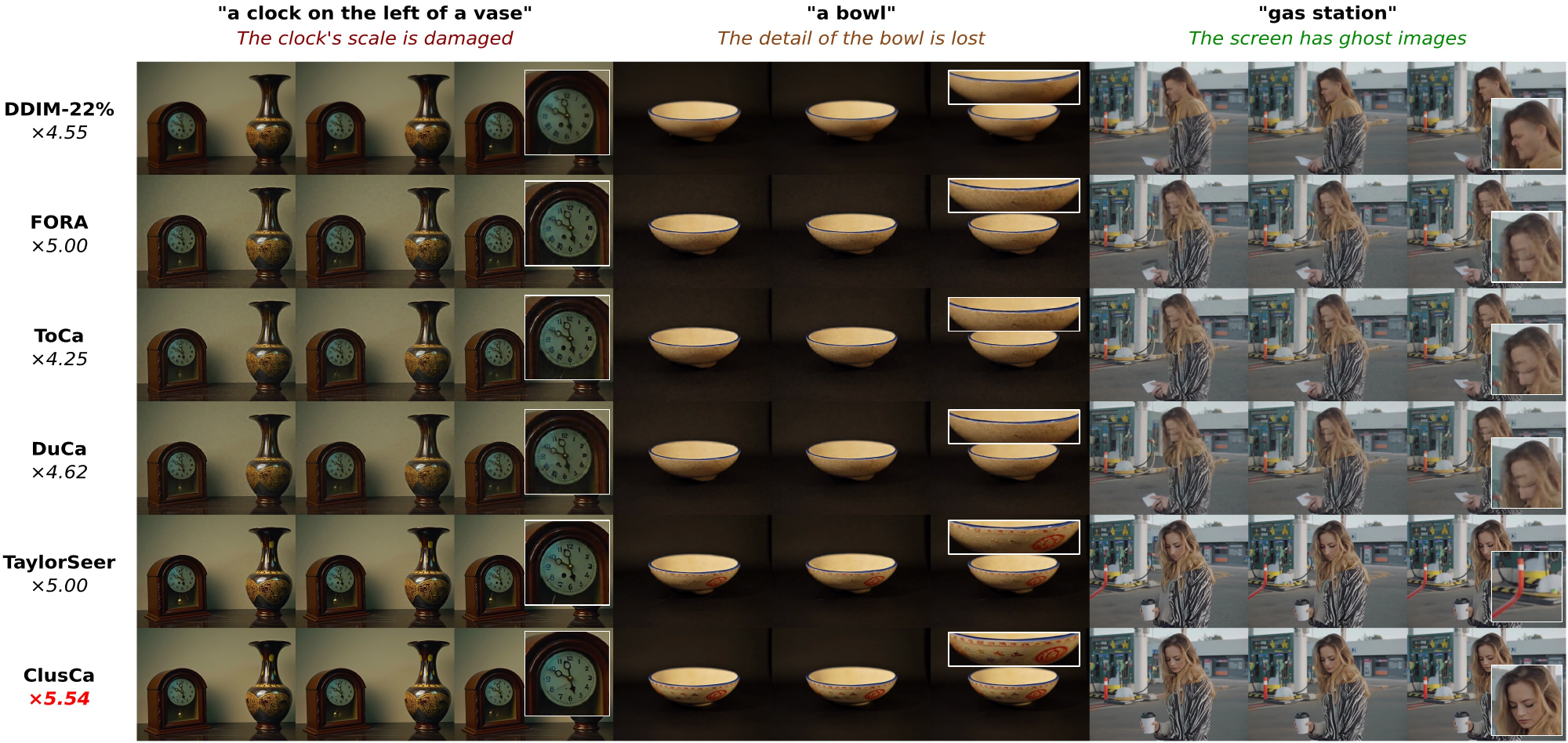}
    \vspace{-2mm}
    \caption{Visualization of different generation methods on various prompts.  ClusCa produces high-quality, accurate images, while other methods suffer from issues such as object damage, missing details, and image ghosting.}
    \label{fig:hunyuan_vislization}
\end{figure*}

As shown in Figure \ref{fig:hunyuan_vislization}, the current acceleration method faces challenges such as object distortion, loss of details, and image ghosting. In contrast, \textit{ClusCa} effectively produces accurate and high-quality video results with more details. In "a clock on the left of a vase" scenario, baseline methods either generate the number "3" as "2" incorrectly, or just a blob of ink, while \textit{ClusCa} generates clearly identifiable digit "3". It also successfully reconstructs intricate surface patterns on the bowl and achieves the balance of portrait and background quality (contrasting with Taylorseer's distorted background objects like unnaturally warped fire hydrants) in the second and third cases respectively. 

\subsection{Results on Class-to-Image Generation}

\begin{table}[ht]

    \centering

    \caption{\textbf{Quantitative comparison in class-to-image generation} on ImageNet with \text{DiT-XL/2.}}
    \setlength\tabcolsep{7pt} 
    \small
    \resizebox{1.02\linewidth}{!}
    {
    \begin{tabular}{l | c | c c c | c c}
\toprule
\bf Method  


& \bf Latency(s) $\downarrow$ & \bf FLOPs(T) $\downarrow$ & \bf Speed $\uparrow$  & \bf FID $\downarrow$ & \bf sFID $\downarrow$  \\
\midrule
{\textbf{$\text{DDIM-25 steps}$}} 
& {0.230}  & {11.87}  & {2.00$\times$}  &  {3.18} &  {4.74}\\
\textbf{FORA} ($\mathcal{N}=2$) 
& 0.278 & 12.35 & {1.92$\times$}  & 2.66 & 4.88\\
\textbf{\texttt{ToCa}} ($\mathcal{N}=3$)  
& {0.216} & {9.73} & {2.44$\times$} & {2.87}   & {4.76}\\
\textbf{\texttt{DuCa}} ($\mathcal{N}=3$) 
& {0.208} & {9.54} & {2.49$\times$} & {2.85}  & {4.64}\\
\rowcolor{gray!20}
$\textbf{ClusCa} $ $(\mathcal{N}=3,K=16)$
& {0.232} & \bf{9.18} & \bf{2.59$\times$} & \bf{2.38}  & \underline{4.76}\\
\rowcolor{gray!20}
$\textbf{ClusCa} $ $(\mathcal{N}=3,K=32)$
& {0.238} & {9.78} & {2.43$\times$} & \underline{2.38}  & \bf{4.69}\\

\midrule     

{\textbf{$\text{DDIM-20 steps}$}} 
& {0.191}  & {9.49}  & {2.50$\times$}  &  {3.81} &  {5.15}\\
\textbf{FORA} ($\mathcal{N}=3$) 
& 0.222 & 8.58 & {2.77$\times$}  & 3.55 & 6.36\\
\textbf{\texttt{ToCa}} ($\mathcal{N}=4$)  
& {0.197} & {8.10} & {2.93$\times$} & {3.42}   & {5.12}\\
\textbf{\texttt{DuCa}} ($\mathcal{N}=4$) 
& {0.175} & {7.61} & {3.12$\times$} & {3.42}  & {4.94}\\
\rowcolor{gray!20}
$\textbf{ClusCa} $ ($\mathcal{N}=4,K=16$)
& {0.189} & \bf{7.35} & \bf{3.23$\times$} & \bf{2.51}  & \underline{5.03}\\

\rowcolor{gray!20}
$\textbf{ClusCa} $ ($\mathcal{N}=4,K=32$)
& {0.202} & {8.03} & {2.95$\times$} & \underline{2.50}  & \bf{4.91}\\

\midrule

{\textbf{$\text{DDIM-15 steps}$}} 
& {0.151}  & {7.12}  & {3.33$\times$}  &  {5.17} &  {6.11}\\
\textbf{FORA} ($\mathcal{N}=4$) 
& 0.193 & 6.66 & {3.56$\times$}  & 4.75 & 8.43\\
\textbf{\texttt{ToCa}} ($\mathcal{N}=5$)  
& {0.176} & {6.77} & {3.51$\times$} & {6.20}   & {7.17}\\
\textbf{\texttt{DuCa}} ($\mathcal{N}=5$) 
& {0.152} & {6.27} & {3.79$\times$} & {6.06}  & {6.72}\\

$\textbf{TaylorSeer}$ $(\mathcal{N}=4,O=1)$
& {0.186}& {6.66} & {3.56$\times$} & {2.71} & {5.45}\\

\rowcolor{gray!20}
$\textbf{ClusCa} $ $(\mathcal{N}=5,K=16)$
& {0.166} & \bf{5.98} & \bf{3.97$\times$} & \bf{2.65}  & \underline{5.13}\\
\rowcolor{gray!20}
$\textbf{ClusCa} $ $(\mathcal{N}=5,K=32)$
& {0.179} & {6.73} & {3.53$\times$} & \underline{2.78}  & \bf{5.01}\\
\midrule 

{\textbf{$\text{DDIM-12 steps}$}} 
& {0.128}  & {5.70}  & {4.16$\times$}  &  {7.80} &  {8.03}\\
\textbf{FORA} ($\mathcal{N}=5$) 
& 0.171 & 5.24 & {4.53$\times$}  & 6.58 & 11.29\\
\textbf{\texttt{ToCa}} ($\mathcal{N}=6$)  
& {0.170} & {6.34} & {3.75$\times$} & {6.55}   & {7.10}\\
\textbf{\texttt{DuCa}} ($\mathcal{N}=6$) 
& {0.145} & {5.81} & {4.08$\times$} & {6.40}  & {6.71}\\


\rowcolor{gray!20}

$\textbf{ClusCa} $ $(\mathcal{N}=6,K=8)$
& {0.153} & \bf{5.15} & \bf{4.61$\times$} & \underline{3.20}  & \underline{5.93}\\
\rowcolor{gray!20}

$\textbf{ClusCa} $ $(\mathcal{N}=6,K=16)$
& {0.159} & {5.52} & {4.29$\times$} & \bf{3.16}  & \bf{5.65}\\

\midrule 

{\textbf{$\text{DDIM-10 steps}$}} 
& {0.112}  & {4.75}  & {5.00$\times$}  &  {12.15} &  {11.33}\\
\textbf{FORA} ($\mathcal{N}=6$) 
& 0.165 & 4.76 & {4.99$\times$}  & 9.15 & 14.84\\
\textbf{\texttt{ToCa}} ($\mathcal{N}=9$)  
& {0.158} & {4.54} & {5.23$\times$} & {12.87}   & {12.82}\\

\textbf{\texttt{DuCa}} ($\mathcal{N}=9$)  
& {0.131} & {4.30} & {5.52$\times$} & {12.05}   & {11.82}\\

$\textbf{TaylorSeer}$ ($\mathcal{N}=6,O=1$) 
& {0.157} & {4.76} & {4.98$\times$} & {3.62} & {7.41}\\

\rowcolor{gray!20}
$\textbf{ClusCa} $ $(\mathcal{N}=7,K=4)$
& {0.133} & \textbf{4.02} & \textbf{5.91$\times$} & \textbf{3.59}  & \underline{6.28}\\
\rowcolor{gray!20}
$\textbf{ClusCa} $ $(\mathcal{N}=7,K=8)$
& {0.138} & {4.21} & {5.63$\times$} & \underline{3.56}  & \textbf{6.05}\\

\bottomrule
\end{tabular}}

\label{table:DiT_Metrics}
    
\end{table}

We compared \textit{ClusCa}  with current cache-based methods such as ToCa~\cite{ToCa},
FORA~\cite{FORA}, DuCa~\cite{DuCa}, and reduced DDIM steps on DiT-
XL/2~\cite{DiT}, demonstrating that \textit{ClusCa} has superior performance than the other methods in both acceleration rate and generation quality. 
\textit{ClusCa} achieves an FID of $2.37$ while providing a $2.72\times$ acceleration, showing that \textit{ClusCa} retains high quality at high acceleration ratios without incurring much computational overhead. At a $4.79\times$ acceleration, our method maintains an FID of $2.96$, outperforming state-of-the-art models such as ToCa and DuCa. 
Notably, as the acceleration ratio increases beyond $3.3\times$, other methods like FORA, ToCa, and DuCa exhibit a significant decline in FID, indicating a collapse in image quality, while \textit{ClusCa} maintains stable performance. This robustness stems from the effectiveness of our approach in choosing important cache features and updating other related tokens. Unlike step-wise cache chosen approaches that accumulate errors under extreme acceleration, our method ensures consistent quality even at extreme speedups.


\begin{figure*}
    \centering
    \begin{minipage}{0.32\linewidth}
        \centering
        \includegraphics[width=0.95\linewidth]{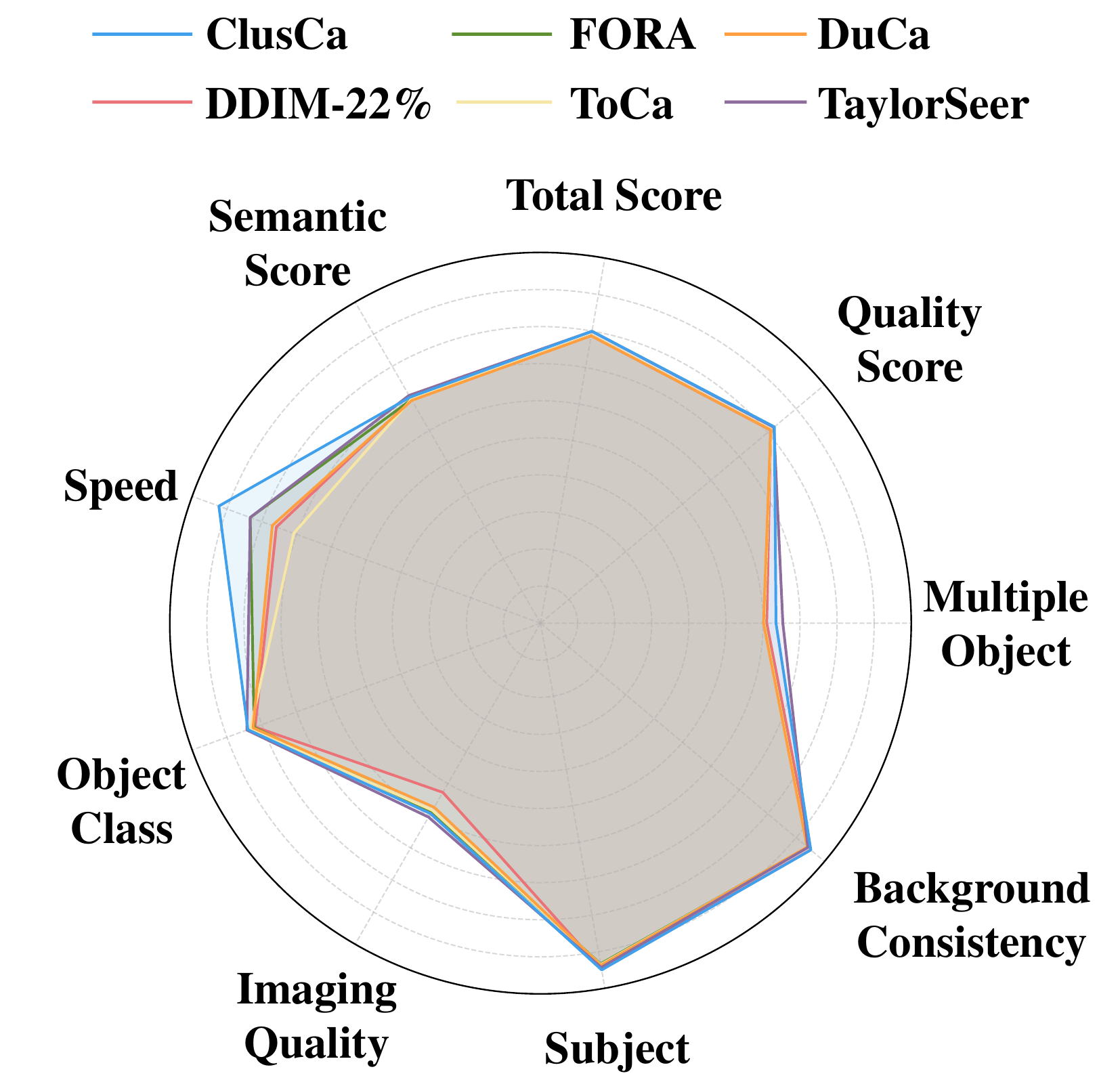}
        \caption{Comparison between ClusCa and other methods on Vbench.}
        \label{fig:VBench-Radar}
    \end{minipage}
    \hfill
    \begin{minipage}{0.24\linewidth}
        \centering
        \includegraphics[width=0.9\linewidth]{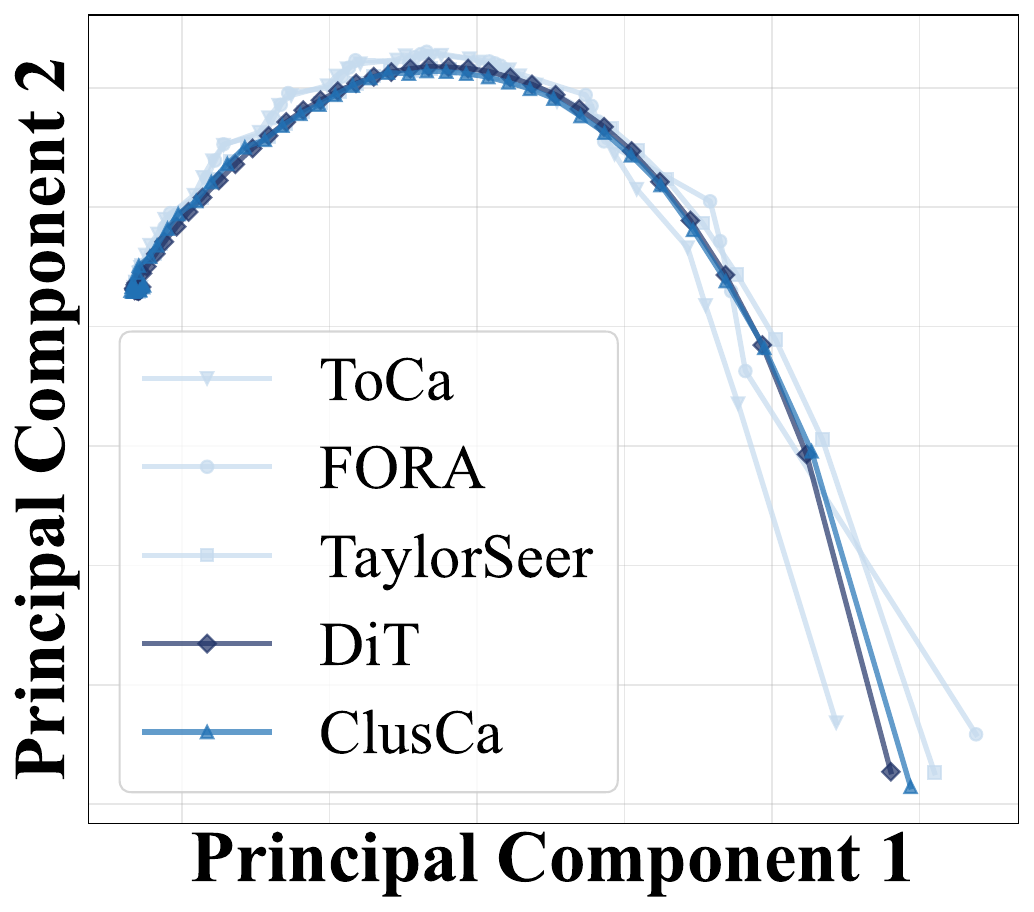}
        \caption{PCA visualization on the trajectories of features in different timesteps. ``DiT'' here denotes the original model without acceleration, which can be considered as the ground-truth trajectory.}
        \label{fig:Trajetories-of-Different-Methods}
    \end{minipage}
    \hfill
    \begin{minipage}{0.40\linewidth}
        \centering
        \includegraphics[width=0.9\linewidth]{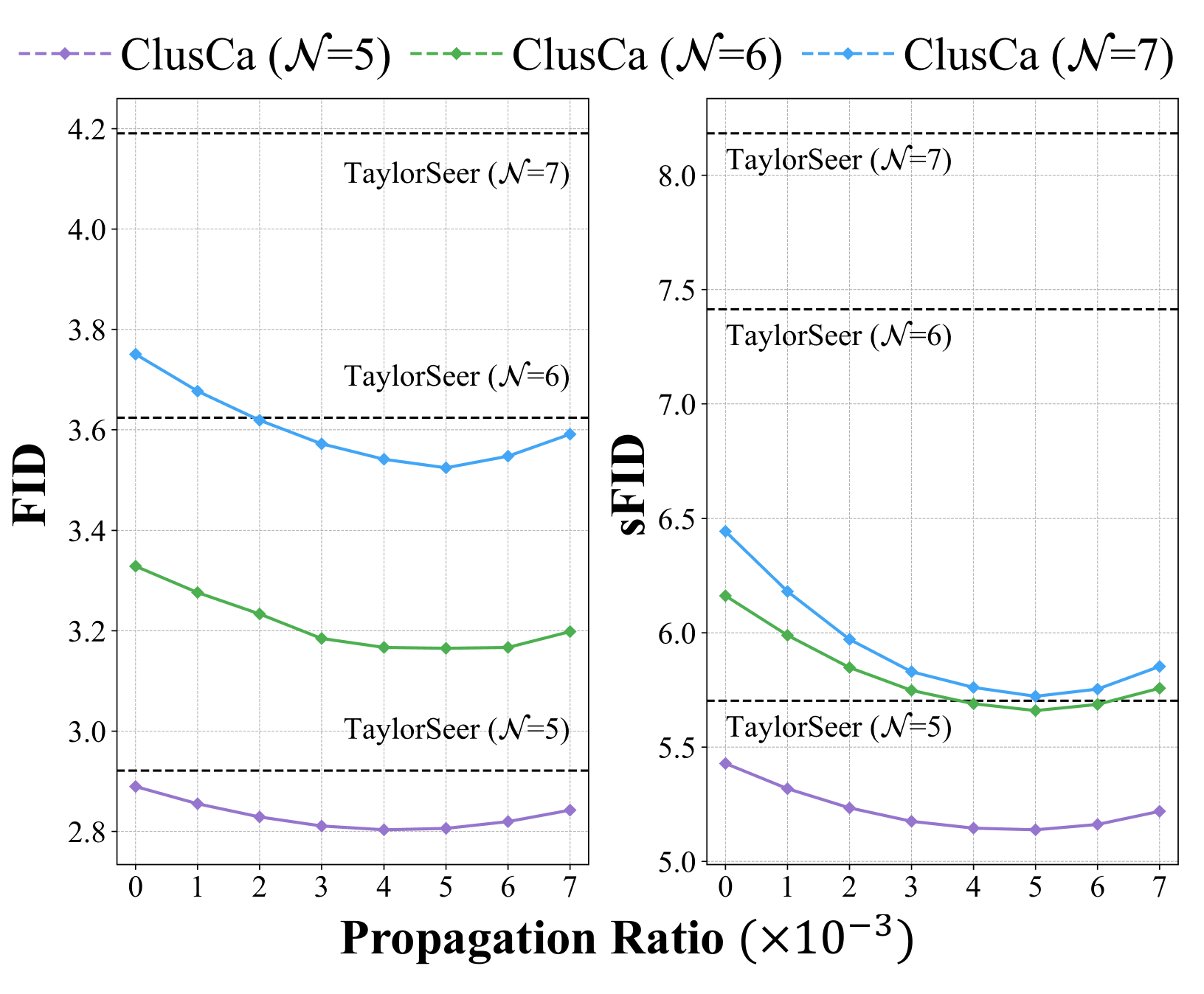}
        \vspace{-4mm}
        \caption{ClusCa with different propagation ratios.}
        \label{fig:propagation-ratio}
    \end{minipage}
    \vspace{-2mm}
\end{figure*}

\subsection{Ablation Study}
Our ablation studies are predominantly conducted on DiT-XL/2, focusing on investigating the impact of the propagation ratio $\gamma$ on generation quality. As hypothesized earlier, the propagation ratio governs the proportional composition of two components in updating non-computed tokens' caches: (1) \emph{spatial reuse:} information from co-clustered recomputed tokens and (2) \emph{temporal reuse:} their historical cached features. Excessively large $\gamma$ values lead to cluster-derived information dominance, effectively degenerating the method into token pruning, while overly small values neglect cluster-wise updates, reducing the approach to ToCa-style caching that only updates computed tokens. Empirical findings in Figure~\ref{fig:propagation-ratio} validate this hypothesis – generation quality demonstrates an initial improvement followed by progressive degradation as $\gamma$ increases. Notably, this quality gain becomes particularly pronounced under large interval configurations: with only an additional computational overhead of $\mathcal{O}(K)$ tokens, the propagation update mechanism achieves up to 0.2 FID reduction, thereby offering a viable solution for maintaining high generation quality under high acceleration ratios. Besides, \textit{ClusCa} achieves consistently lower FID than the previous SOTA TaylorSeer, demonstrating its effectiveness.

\vspace{-1em}

\subsection{Visualization on Feature Trajectories}
We perform principal component analysis (PCA) on all features from a specific layer across different timesteps on DiT-XL/2 in Figure~\ref{fig:Trajetories-of-Different-Methods}, projecting them into a 2D latent space for visualization. The trajectory denoted by ``DiT'' indicates the original DiT without acceleration, while the other trajectories correspond to other caching-based acceleration methods. It is observed that our method achieves the highest degree of matching with the original unaccelerated DiT's feature evolution patterns among all compared approaches, especially at the end of the trajectory, quantitatively verifying that \textit{ClusCa} successfully minimizes error accumulation while maintaining good computational efficiency.




\section{Conclusion}\label{sec:conclusion}
Motivated by significant similarity in spatial dimension, we introduce \textit{ClusCa}, an efficient feature caching framework that accelerates diffusion models by jointly exploiting spatial and temporal token similarities. By selectively computing as few as 16 tokens at intermediate denoising steps, \textit{ClusCa} efficiently propagates feature updates across all tokens, mitigating error accumulation caused by prolonged caching cycles. Experimental results demonstrate that \textit{ClusCa} achieves significant quality improvements (e.g. 0.9949 Image Reward at 4.5× speedup) while maintaining competitive acceleration performance, effectively improving the efficiency-quality trade-off inherent to existing feature caching approaches, providing insight for leveraging both spatial and temporal similarity in diffusion transformer acceleration. 

\begin{acks}
This work was partially supported by Dream Set Off - Kunpeng\&Ascend Seed Program. 
\end{acks}

\bibliographystyle{ACM-Reference-Format}
\balance
\bibliography{main}

\appendix
\clearpage
\setcounter{page}{1}

\section{Appendix Overview}
This appendix provides supplementary information to support the main paper's findings, focusing on experimental details, in-depth ablation studies, additional results, and a computational cost analysis for the proposed method, \textit{ClusCa}.
\begin{itemize}
    \item \textbf{Section B:} Experimental details of model configurations and hardware setups.
    \item \textbf{Section C:} Supplementary results for ablation studies on key hyperparameters ($\gamma$, $K$, $\mathcal{N}$).
    \item \textbf{Section D:} Additional evaluation results on video dynamics and fidelity.
    \item \textbf{Section E:} Analysis of the computational overhead of the clustering and propagation mechanisms.
\end{itemize}

\section{Experiment Details}
In this section, we detail our experiments mentioned in section 4.1. 
\begin{itemize}
    \item \textbf{FLUX.1-dev} employs a 50-step Rectified Flow sampling strategy as its core generation mechanism. We place the clustering and propagation mechanisms in double-stream blocks to select and partially compute img and txt tokens. Clipscore is computed using laion/CLIP-ViT-g-14-laion2B-s12B-b42K~\cite{ilharco_gabriel_2021_5143773} pretrained model. All evaluations were conducted on NVIDIA A800 GPUs with 80GB memory, focusing on baseline performance metrics and generation quality.
    \item \textbf{Hunyuan-Video} ~\cite{li2024hunyuanditpowerfulmultiresolutiondiffusion,sun2024hunyuanlargeopensourcemoemodel} is benchmarked under its native HunyuanLarge architecture, adhering to a standardized 50-step inference protocol. The model retains default hyperparameters for sampling consistency while incorporating the activation period $\mathcal{N}$ and cluster number $K$. Latency profiling was performed on NVIDIA H20 96GB GPUs, with full-scale inference tasks executed on H100 80GB GPUs to ensure precise resource allocation and runtime accuracy. In practice, we find that the last timestep is crucial in Video Consistency. Without activating the last timestep, the generated videos will produce visible temporal flickering, which affects overall video quality. Inspired by previous work\cite{ToCa}, we rearranged some activation steps, maintaining the total number of activation steps unchanged, so that the last timestep can be activated. 
    \item \textbf{DiT-XL/2} utilizes a 50-step DDIM sampler aligned with the comparative framework of other models.  All the experiments are conducted on NVIDIA A800 GPUs with 80GB memory. 
\end{itemize}

\section{Supplementary Results for Ablation Studies}
\subsection{Ablation on propagation ratio $\gamma$}
To systematically investigate the impact of the propagation ratio — a critical hyperparameter governing the composition balance within the non-computational token cache — we conducted extensive ablation studies on the Diffusion Transformer (DiT) architecture. As illustrated in Table~\ref{table:propagation-ratio} (with fixed hyper-parameters $\mathcal{O}=1,K=16$), all configurations with same interval $\mathcal{N}$ maintain identical computational complexity (FLOPs) and speedup ratios, while exclusively varying propagation ratios from 0.0 to 0.007. The observed performance variations under equivalent computational budgets demonstrate the efficacy of our cluster-aware propagation mechanism: improves the quality without increasing any extra computational cost. 
\begin{table}[htb]
\centering
\caption{\textbf{Ablation Study with Different Configurations} on ImageNet with \text{DiT-XL/2.}}
\label{table:propagation-ratio}
\begin{tabularx}{\linewidth}{c|c|>{\centering\arraybackslash}X >{\centering\arraybackslash}X}
\toprule
\multicolumn{2}{c|}{\textbf{Configuration}} & \bf sFID$\downarrow$ & \bf FID$\downarrow$ \\
\midrule


\multirow{8}{*}{\textbf{$\mathcal{N}=6$}}
& $\gamma=0.000$ & 6.161 & 3.328 \\
& $\gamma=0.001$ & 5.989 & 3.276 \\
& $\gamma=0.002$ & 5.848 & 3.233 \\
& $\gamma=0.003$ & 5.748 & 3.184 \\
& $\gamma=0.004$ & 5.689 & 3.166 \\
& $\gamma=0.005$ & \textbf{5.659} & \textbf{3.165} \\
& $\gamma=0.006$ & 5.686 & 3.166 \\
& $\gamma=0.007$ & 5.757 & 3.198 \\
\midrule

\multirow{8}{*}{\textbf{$\mathcal{N}=7$}}
& $\gamma=0.000$ & 6.444 & 3.751 \\
& $\gamma=0.001$ & 6.181 & 3.677 \\
& $\gamma=0.002$ & 5.971 & 3.619 \\
& $\gamma=0.003$ & 5.830 & 3.572 \\
& $\gamma=0.004$ & 5.761 & 3.541 \\
& $\gamma=0.005$ & \textbf{5.722} & \textbf{3.524} \\
& $\gamma=0.006$ & 5.754 & 3.547 \\
& $\gamma=0.007$ & 5.853 & 3.591 \\
\midrule

\multirow{8}{*}{\textbf{$\mathcal{N}=8$}}
& $\gamma=0.000$ & 6.775 & 4.782 \\
& $\gamma=0.001$ & 6.452 & 4.659 \\
& $\gamma=0.002$ & 6.192 & 4.557 \\
& $\gamma=0.003$ & 6.004 & 4.467 \\
& $\gamma=0.004$ & 5.877 & 4.401 \\
& $\gamma=0.005$ & \textbf{5.830} & \textbf{4.386} \\
& $\gamma=0.006$ & 5.855 & 4.412 \\
& $\gamma=0.007$ & 5.986 & 4.439 \\
\bottomrule
\end{tabularx}
\end{table}

\subsection{Ablation on number of clusters K}
Regarding the number of clusters, K, a larger value indeed improves generation quality. This is because more tokens are recomputed, and smaller clusters exhibit significantly higher intra-cluster similarity. However, this comes at the cost of increased computational load. After balancing this trade-off between fidelity and acceleration, we ultimately selected K=16 for our experiments on DiT-XL/2.
\begin{table}[htb]

\caption{\textbf{Ablation study on number of clusters K.}}

\begin{tabular}{l|ll|ll}
\toprule
\multicolumn{1}{c}{Method} & \multicolumn{1}{c}{FLOPs}                           & \multicolumn{1}{c}{Speedup}                         & \multicolumn{1}{c}{FID}                              & \multicolumn{1}{c}{sFID}                             \\

\midrule

ClusCa(K = 4)              & {\color[HTML]{333333} 5.42}                         & {\color[HTML]{333333} 4.37}                         & {\color[HTML]{333333} 2.819}                         & {\color[HTML]{333333} 5.285}                         \\
ClusCa(K = 8)              & \cellcolor[HTML]{F8F8F8}{\color[HTML]{333333} 5.61} & \cellcolor[HTML]{F8F8F8}{\color[HTML]{333333} 4.23} & \cellcolor[HTML]{F8F8F8}{\color[HTML]{333333} 2.806} & \cellcolor[HTML]{F8F8F8}{\color[HTML]{333333} 5.226} \\
ClusCa(K = 16)             & {\color[HTML]{333333} 5.98}                         & {\color[HTML]{333333} 3.96}                         & {\color[HTML]{333333} 2.803}                         & {\color[HTML]{333333} 5.145}                         \\
ClusCa(K = 32)             & \cellcolor[HTML]{F8F8F8}{\color[HTML]{333333} 6.72} & \cellcolor[HTML]{F8F8F8}{\color[HTML]{333333} 3.53} & \cellcolor[HTML]{F8F8F8}{\color[HTML]{333333} 2.784} & \cellcolor[HTML]{F8F8F8}{\color[HTML]{333333} 5.006}\\

\bottomrule

\end{tabular}

\label{table:ablation-k}

\end{table}

\begin{table*}[htbp]
\centering
\caption{\textbf{Comparison of additional fidelity scores in video generation(HunyuanVideo).} }
\tiny
\renewcommand{\arraystretch}{1.2} 
\setlength\tabcolsep{7.0pt}
\resizebox{0.9\linewidth}{!}{
\begin{tabular}{l|cc|c|ccc}
\toprule
{\bf Method} & \multicolumn{2}{c|}{\bf Acceleration} & {\bf VBench $\uparrow$}  & \multirow{2}{*}{\bf PSNR$\uparrow$} & \multirow{2}{*}{\bf SSIM$\uparrow$} & \multirow{2}{*}{\bf LPIPS$\downarrow$}\\
\cline{2-3}
{\bf HunyuanVideo} & {\bf FLOPs(T) $\downarrow$} & {\bf Speed $\uparrow$} & \bf Score(\%)  & & & \\
\midrule
$\textbf{Original: 50 steps}$ & 29773.0 & 1.00$\times$ & 80.66 & - & - & -\\ 
\midrule

    $\textbf{DDIM-22\%}$  
                            & {6550.1} & {4.55$\times$} & {78.74} & 11.2576 &	0.4323	& 0.6473\\
    $\textbf{FORA}$ $(\mathcal{N}=5)$ 
                            & {5960.4} & {5.00$\times$} & {78.83} &11.6969&0.4343&0.6354\\
    $\textbf{\texttt{ToCa}}$ $(\mathcal{N}=5)$ 
                            & {7006.2} & {4.25$\times$} & {78.86} &11.3269	&0.4243& 0.6468\\
    $\textbf{\texttt{DuCa}}$ $(\mathcal{N}=5)$ 
                            & {6483.2} & {4.62$\times$} & {78.72} &11.3366&0.4279& 0.6467\\
    \textbf{TeaCache}  $({l}=0.4)$
                            & 6550.1 & 4.55$\times$ & 79.36 & 11.6859&0.4305& 0.6177\\
    
    $\textbf{TaylorSeer}$ $(\mathcal{N}=5,O=1)$ 
                            & {5960.4} & {5.00$\times$} & {79.93} &11.5505&0.4238&0.6303\\
    

    \rowcolor{gray!20}
    $\textbf{ClusCa}$ $(\mathcal{N}=5,K=32)$ 
                            & {5968.1} & {4.99$\times$} & {\textbf{79.99}} &\textbf{13.5328}&\textbf{0.4863}&\textbf{0.5413}\\
    
    \rowcolor{gray!20}
    $\textbf{ClusCa}$ $(\mathcal{N}=6,K=32)$ 
                            & {5373.0} & {{5.54}$\times$} & \underline{79.96} &\underline{12.3300}&\underline{0.4533}&\underline{0.5957}\\

    \bottomrule

\end{tabular}}
\label{table:HunyuanVideo-More}
\end{table*}

\begin{table*}[htbp]
\centering
\caption{\textbf{Comparison of VBench motion sub-metrics in video generation (HunyuanVideo).}}
\tiny
\renewcommand{\arraystretch}{1.2}
\setlength\tabcolsep{2.1pt}
\resizebox{0.90\linewidth}{!}{%
\begin{tabular}{l|c|ccccccc|c}
\toprule
\textbf{Method} & \textbf{Speed} & \textbf{BG Cons.} & \textbf{Temp. Flick.} & \textbf{Motion Smooth.} & \textbf{Dyn. Deg.} & \textbf{Human Act.} & \textbf{Temp. Style} & \textbf{Overall Cons.} & \textbf{Avg.} \\
\midrule
\textbf{FORA}$(\mathcal{N}=5)$        & 5.00$\times$ & 0.9403 & 0.9771 & 0.9688 & 0.2764 & 0.907 & 0.6602 & 0.7209 & 0.7787 \\
\textbf{ToCa}$(\mathcal{N}=5)$        & 4.25$\times$ & 0.9399 & 0.9781 & 0.9708 & 0.2778 & 0.898 & 0.6615 & 0.7217 & 0.7783 \\
\textbf{DuCa}$(\mathcal{N}=5)$        & 4.62$\times$ & 0.9393 & 0.9798 & 0.9736 & 0.2778 & 0.905 & 0.6613 & 0.7184 & 0.7793 \\
\textbf{Taylor}$(\mathcal{N}=5)$    & 5.00$\times$ & 0.9416 & 0.9819 & 0.9739 & 0.2986 & 0.914 & 0.6720 & 0.7293 & 0.7873 \\
\rowcolor{gray!20}
\textbf{ClusCa}$(\mathcal{N}=6,K=32)$     & 5.54$\times$ & 0.9518 & 0.9865 & 0.9787 & 0.3000 & 0.914 & 0.6706 & 0.7254 & 0.7896 \\
\bottomrule
\end{tabular}%
}
\label{table:Hunyuan-VBench}
\end{table*}

\subsection{Ablation on intervals N}
As for the cache interval, N, a larger interval inevitably increases the error between the cached and true features. Our method effectively mitigates the quality degradation from longer intervals. As demonstrated in Table 3, our approach with N=7 and N=5 still outperforms TaylorSeer baseline using shorter intervals of N=6 and N=4, respectively, showcasing its robustness.
\begin{table}[htb]

\caption{\textbf{Ablation study on cache cycle length N.}}

\begin{tabular}{l|ll|ll}
\toprule
\multicolumn{1}{c}{Method}         & FLOPs                       & Speedup                     & FID                         & sFID                        \\

\midrule

{\color[HTML]{333333} ClusCa(N=3)} & {\color[HTML]{333333} 9.17} & {\color[HTML]{333333} 2.59} & {\color[HTML]{333333} 2.40} & {\color[HTML]{333333} 4.70} \\
\rowcolor[HTML]{F8F8F8} 
{\color[HTML]{333333} ClusCa(N=4)} & {\color[HTML]{333333} 7.35} & {\color[HTML]{333333} 3.23} & {\color[HTML]{333333} 2.51} & {\color[HTML]{333333} 5.03} \\
{\color[HTML]{333333} ClusCa(N=5)} & {\color[HTML]{333333} 5.98} & {\color[HTML]{333333} 3.96} & {\color[HTML]{333333} 2.80} & {\color[HTML]{333333} 5.14} \\
\rowcolor[HTML]{F8F8F8} 
{\color[HTML]{333333} ClusCa(N=6)} & {\color[HTML]{333333} 5.52} & {\color[HTML]{333333} 4.29} & {\color[HTML]{333333} 3.17} & {\color[HTML]{333333} 5.66} \\
{\color[HTML]{333333} ClusCa(N=7)} & {\color[HTML]{333333} 4.62} & {\color[HTML]{333333} 5.14} & {\color[HTML]{333333} 3.52} & {\color[HTML]{333333} 5.72}\\

\bottomrule
\end{tabular}

\label{table:ablation-N}

\end{table}

\section{Dynamics on video and additional fidelity measures.}
We report the detailed sub-metrics of VBench motion-related score in Table \ref{table:HunyuanVideo-More}, and evaluate these methods in fidelity measure (see Table \ref{table:Hunyuan-VBench}). As shown in the tables, ClusCa achieves remarkable performance, maintaining high-quality motion scores at a significantly reduced computational cost. Notably, at the highest acceleration ratio of $5.54\times$, ClusCa still delivers the best fidelity across all key metrics—achieving the highest PSNR (\textbf{13.53}), the highest SSIM (\textbf{0.4863}), and the lowest LPIPS (\textbf{0.5413}). This demonstrates that our method not only improves generation speed but also effectively preserves both structural integrity and perceptual realism of the video content, outperforming all compared baselines.

\section{Time Cost Analysis}
We quantitatively analyze the computational overhead of the clustering and propagation mechanisms in the denoising process. Since clustering operations are performed on full computational steps, they introduce only marginal computational overhead. The propagation mechanism incurs slightly higher costs, which aligns with the inherent characteristic of token-wise feature update methods like ToCa~\cite{ToCa} and DuCa~\cite{DuCa} that require partial feature updates. Although introducing approximately 10\% additional computational overhead, this design notably enhances the generation quality. As demonstrated in Figure 2 in the main text, \textit{ClusCa} achieves superior balance between inference speed and generation quality compared to existing methods. 

\begin{figure}[htp]
    \centering
    \includegraphics[width=0.98\linewidth]{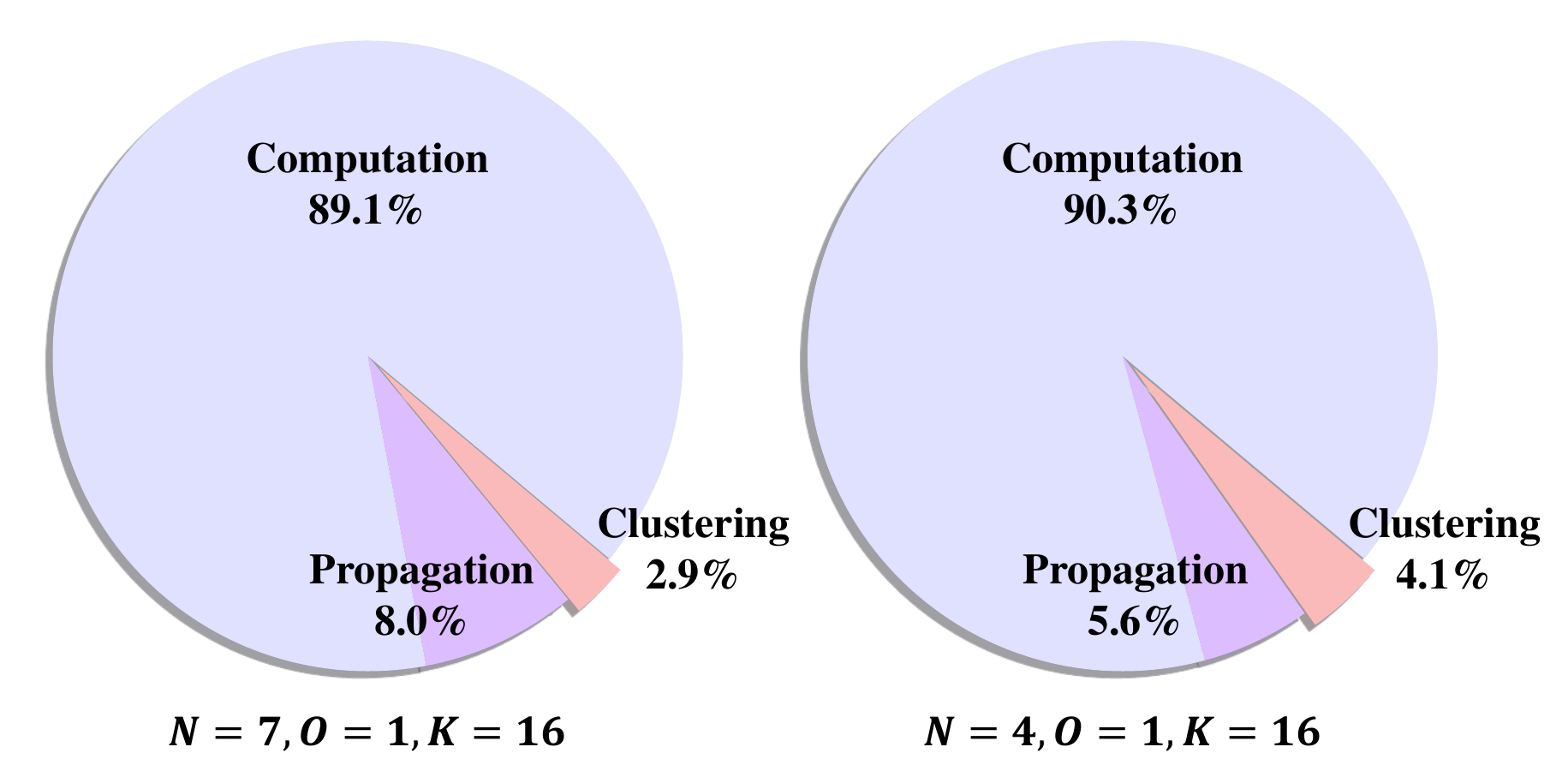}
    \caption{Time consumption visualization of each part}
    \label{fig:Time consumption visualization}
\end{figure}

\end{document}